\documentclass{article}

\PassOptionsToPackage{numbers, compress}{natbib}

    \usepackage[preprint]{neurips_2021}

\usepackage[utf8]{inputenc} 
\usepackage[T1]{fontenc}    
\usepackage{hyperref}       
\usepackage{url}            
\usepackage{booktabs}       
\usepackage{amsfonts}       
\usepackage{nicefrac}       
\usepackage{microtype}      
\usepackage{xcolor}         
\usepackage{graphicx}
\usepackage{array}
\usepackage{float}
\newcolumntype{P}[1]{>{\centering\arraybackslash}p{#1}}
\newcolumntype{M}[1]{>{\centering\arraybackslash}m{#1}}
\title{A Studious Approach to Semi-Supervised Learning}

\author{Sahil Khose\thanks{Authors have contributed equally to this work and share first authorship}  \quad Shruti Jain$^{*}$  \quad V Manushree$^{*}$ \\
Manipal Institute of Technology, Manipal \\
{\tt\small \{sahil.khose, shruti.jain, manushree.v\}@learner.manipal.edu}
}

\begin{document}
\maketitle

\begin{abstract}%
The problem of learning from few labeled examples while using large amounts of unlabeled data has been approached by various semi-supervised methods. Although these methods can achieve superior performance, the models are often not deployable due to the large number of parameters. This paper is an ablation study of distillation in a semi-supervised setting, which not just reduces the number of parameters of the model but can achieve this while improving the performance over the baseline supervised model and making it better at generalizing. After the supervised pretraining, the network is used as a teacher model, and a student network is trained over the soft labels that the teacher model generates over the entire unlabeled data. We find that the fewer the labels, the more this approach benefits from a smaller student network. This brings forward the potential of distillation as an effective solution to enhance performance in semi-supervised computer vision tasks while maintaining deployability. 
\end{abstract}


\section{Introduction}
\label{section: Introduction}
Deep learning has achieved remarkable success in many visual and linguistic tasks as a result of recent research advancements. On computer vision applications like classification, larger, deeper models trained in a supervised learning framework have produced the state of the art results. Despite their superior performance, these models are not feasible for real-time deployment due to their computational and memory requirements and the unavailability of large labeled datasets.

In most deep learning problems; more parameters, larger datasets, and more compute results in better accuracy. This is a result of the model's capacity to learn more complex functions, thereby increasing the performance. Distillation with KL Divergence Loss is one of the ways being actively explored for transferring this information obtained by these models to much smaller models. It also acts as an excellent regularizer, especially when there is less labeled data.

This paper is an empirical study of distillation based semi-supervised learning to overcome overfitting, a common problem in semi-supervised setup and bettering performance when limited with small deployable models. We experimented using three architectures: Efficient Net-b5 \cite{effnet}, ResNet-18 \cite{resnet}, and MobileNet-V3-Large \cite{mobilenet} to demonstrate the benefit of model compression and four types of label split, highlighting the semi-supervised advantage and model optimization.

\section{Related Work}
\label{section: related work}
Some of the early works that influenced distillation were by Bucila et al. \cite{bucilu}, who used a single neural network that learns by trying to mimic the output of an ensemble of models. The work by Ba and Caruana \cite{deepnets} compresses larger and complex ensembles into small, faster models without much loss in performance using logits. Hinton et al. \cite{hinton_kd} introduced distillation of knowledge in neural networks using the soft target predictions of the teacher model to train the student network. Intelligent teachers that provide additional privileged information to the students to accelerate the learning process were introduced by \cite{lupi}.

There has been recent advanced research exploring distillation, such as Fitnet \cite{fitnet}, where the student learns by mimicking the feature maps of the teacher instead of using the output distribution.  Net2Net \cite{net2net} uses function-preserving transformations to accelerate the transfer of knowledge from smaller neural networks to significantly larger ones. The paper \cite{deep_model_compression} proposes a noise-based regularizer to improve the performance of the student network guided by the teacher. The paper \cite{neuron_selectivity} introduces a novel knowledge transfer method in which the distributions of neuron selectivity patterns are matched between the teacher and student models by minimizing the Maximum Mean Discrepancy (MMD) between them. These works have paved the way for enhancing knowledge transfer in neural networks and help in model compression for real-time deployability.

\section{Methodology}
\label{section: methodology}
In knowledge distillation \cite{hinton_kd}, a model is trained first on the dataset; then it is used as the teacher to transfer its knowledge to another model, the student. This information transmission is accomplished through training students on probability distributions of the teacher's predictions rather than hard labels. Instead of the Cross-Entropy Loss, Kullback–Leibler (KL) Divergence Loss is employed. It is a measure of distance between continuous distributions, in this case, the probability distributions of the teacher and student predictions. The KL Divergence Loss \ref{eqn:kl_loss} for two distributions, P and Q, is calculated as follows:

\begin{equation}
        KL\left(P\ ||\ Q\right)= -\sum P\left (x \right) * \log\left (Q \left (x\right ) / P \left (x \right)\right)
\label{eqn:kl_loss}
\end{equation}

The probability distribution of the output classes represents the measure of the teacher's uncertainty about the prediction, providing additional information to guide the student. To compress the models, we perform knowledge distillation, where the teacher models are larger than the student models. In the case of self-distillation, the student model and the teacher model are the same architecture. Here, we experiment on both knowledge distillation and self-distillation in a semi-supervised setting to obtain better performing and more generalized models.

We use the CIFAR-10 dataset that contains $10$ classes of natural images with a total of $50000$ training and $10000$ testing samples, each of which is a $32$×$32$ RGB image. We preprocessed the data by normalizing and augmenting it by using random crops and horizontal flips. The CIFAR-$10$ dataset is first divided into training, validation, and test sets in the ratio $4$:$1$:$1$, respectively. Our training data is further divided into two categories: labeled data ($X_{lab}$) for which we have labels ($Y_{lab}$) available and unlabeled data ($X_{unlab}$) whose labels ($Y_{unlab}$) will not be used for training the teacher model. We will, however, use the entire training dataset for evaluating the student network.

We have used three models for our study: MobileNetV$3$-Large \cite{mobilenet}, ResNet-$18$ \cite{resnet} and Efficient Net-b$5$ \cite{effnet}. Efficient Net-b$5$ contains roughly 28 million parameters, while ResNet-$18$ has approximately 11 million parameters, one-third of the former. MobileNetV$3$-Large has about 4 million, half the times of ResNet-$18$ and one-seventh of Efficient Net-b$5$. All the models used were pretrained on ImageNet \cite{imagenet}. We perform knowledge distillation from Efficient Net to ResNet and MobileNet; and from ResNet to MobileNet. Along with this, we perform self-distillation on all three models. 
We experiment on three different label percentages: $10$, $25$, $50$ for a semi-supervised approach, along with $100$ percentage labels as the supervised benchmark for comparison. 

\paragraph{Implementation details:}The teacher model on the labeled data ($X_{lab}$, $Y_{lab}$) is trained for $30$ epochs. We use Stochastic Gradient Descent for all three models with a learning rate of $0.001$ for MobileNet and $0.01$ for Efficient Net and ResNet; and momentum of $0.9$ with weight decay of $5e-4$. The criterion for training the teacher is Cross-Entropy Loss.

The entire dataset ($X_{lab}$, $X_{unlab}$) and the probability distributions of the teacher outputs are passed to the student model. We then train the student model by minimizing the distance between the probability distributions of both the teacher and student model using KL Divergence Loss for $30$ epochs. The same optimizer and learning rate as the teacher model are used.

\section{Experiments and Evaluation}
\label{section: evaluation}

\begin{figure}[h]
    \centering
    \includegraphics[scale=0.55]{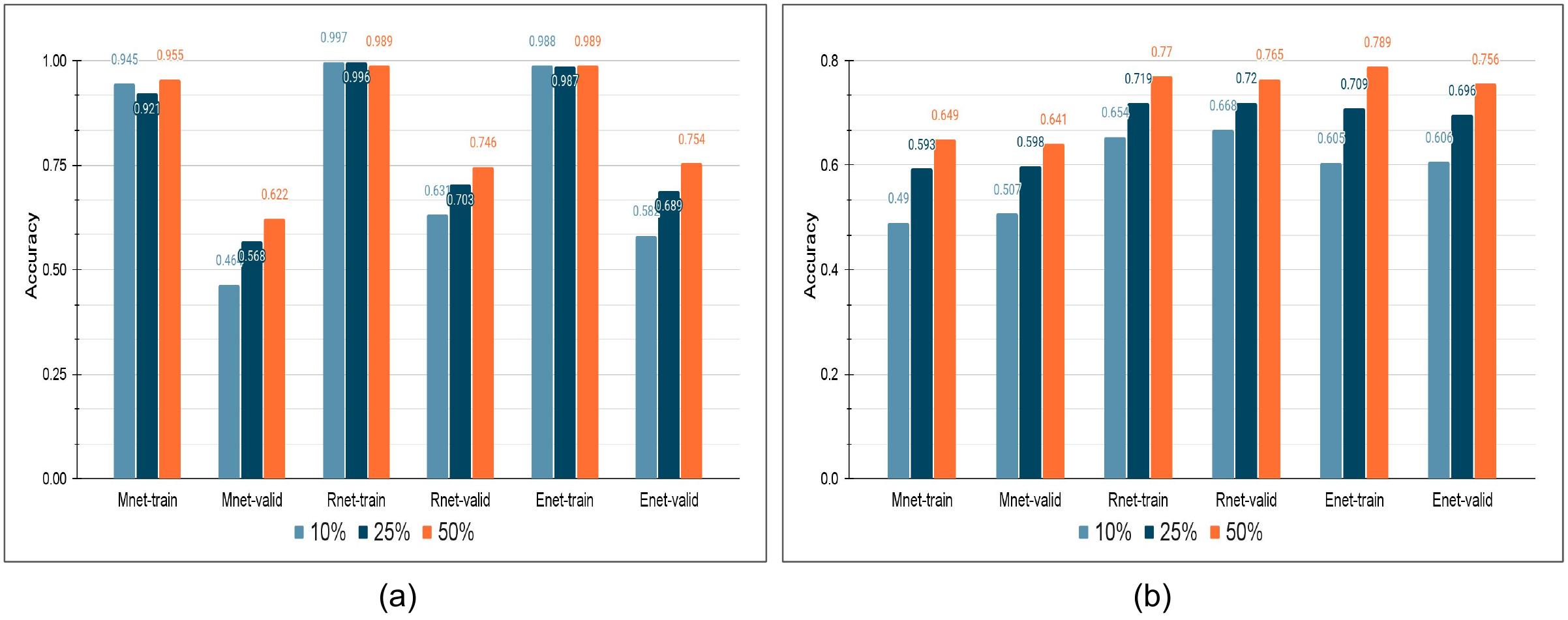}
    \caption{Model Accuracies on different splits before distillation (fig. a), after distillation (fig. b).
    'Mnet', 'Rnet', 'Enet' represents MobileNetV$3$-Large,  ResNet-$18$, Efficient Net-b$5$.}
    \label{fig: accuracies vs splits}
\end{figure}

The performance of teacher models trained on different splits of data are shown in Figure \ref{fig: accuracies vs splits}(a). It is observed that the teacher models are overfitting on all splits displaying substantial variations in the training and validation accuracy, with the larger models showing higher accuracy. We explore knowledge distillation of ResNet to MobileNet; and Efficient Net to ResNet and MobileNet to demonstrate that the models can be compressed with little to no harm to the performance. We further explore self-distillation to show that distillation also acts as a regularizer, as shown in \cite{hinton_kd}.

\subsection{Self-distillation in semi-supervised setup}

\paragraph{Decrease in overfitting:}It is observed that self-distilling the models using KL Divergence Loss enhanced the validation accuracy while simultaneously acting as a regularizer, resulting in a model that generalizes well.  The difference in training and validation accuracy has decreased considerably for self-distilled models, as shown in Figure \ref{fig: accuracies vs splits}(b).

The training accuracy for MobileNet on $10$\% labels dropped from $0.9450$ to $0.4904$, and its validation accuracy increased from $0.4638$ to $0.5073$. Similar behavior can be observed for the other models and other data splits, which is summarized in Figure \ref{fig: accuracies vs splits}(b). 

This is because the student learns richer information from the teacher's uncertainties while making a prediction. As the student attempts to mimic the behavior of the teacher model, it learns not just the correct label for an example but also the probability that the teacher assigns to the other classes.  Since the teacher model is not perfect and overfits the dataset, it adds noise to the dataset and helps in the generalization of the student model.

\begin{figure}[h]
    \centering
    \includegraphics[scale=0.6]{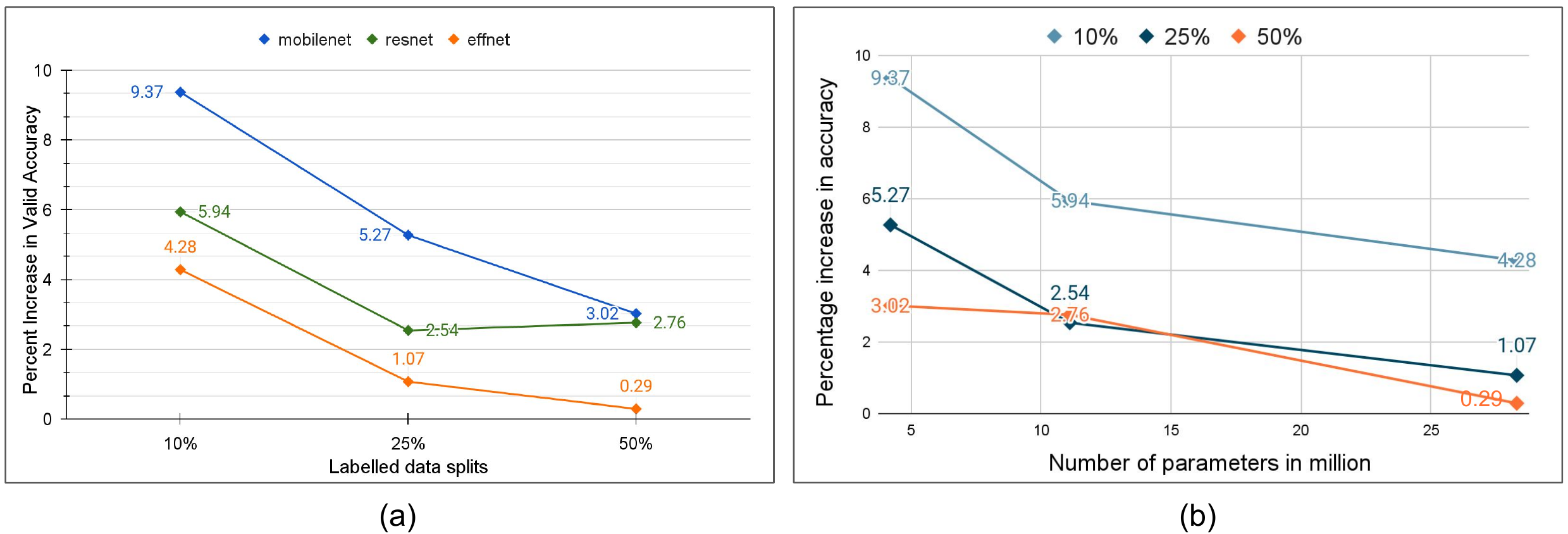}
    \caption{Percentage increase in valid accuracy after self-distilling the models on different splits.}
    \label{fig: percent increase}
\end{figure}

\paragraph{Increase in accuracy:}
It is observed that students that learned from teachers and trained on data splits with lesser labels had greater improvement in validation accuracy. MobileNet had a $9.37$\% increase in validation accuracy compared to the teacher model on $10$\% labeled data and had a $3.02$\% increase with $50$\% labeled data. Another inference drawn from the experiments is that models with lesser parameters showed more significant gains in validation accuracy as compared to the larger models. The inference, lesser the labeled data, and smaller the model; greater is the increase in accuracy during distillation is summarized in the Figure \ref{fig: percent increase} (a) and (b) respectively.

This is due to the fact that when the teacher has more labels or has more parameters, it is more confident in its predictions, which results in the soft labels being similar to the hard labels. As a result, the KL Divergence Loss yields fewer benefits since there is comparatively less information to learn from the probability distribution.

\subsection{Knowledge distillation in semi-supervised setup}
\begin{figure}[h]
    \centering
    \includegraphics[scale=0.6]{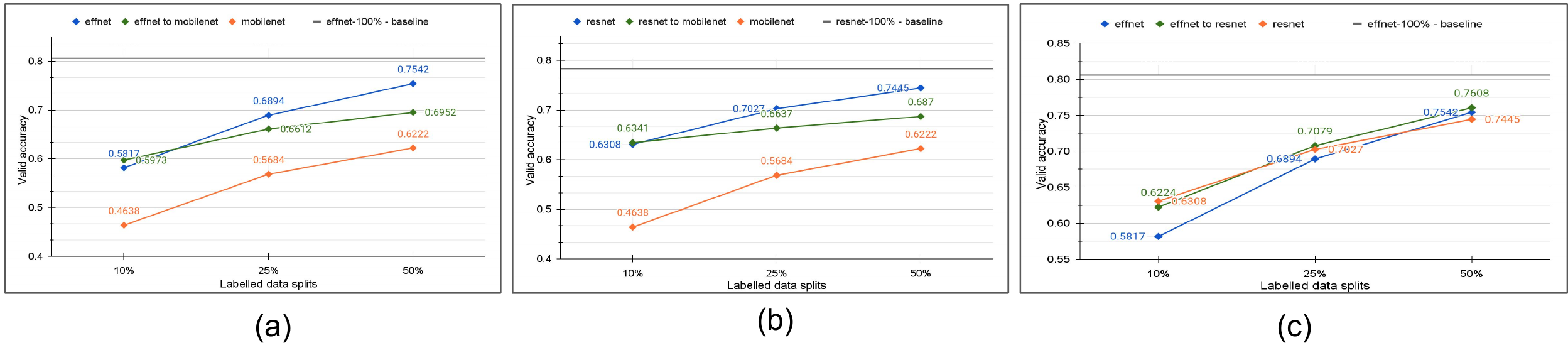}
    \caption{Knowledge distillation of Efficient Net to MobileNet (fig. a), ResNet to MobileNet (fig. b), Efficient Net to ResNet (fig. c).}
    
    \label{fig:knoeledge distill}
\end{figure}

Out of the three models we use, Efficient Net has the largest number of parameters, followed by ResNet and MobileNet. The smaller models can learn and gain knowledge from the bigger models and perform in a similar manner. There is a substantial improvement in accuracy compared to their undistilled counterpart trained directly using hard labels. One of the most significant advantages of knowledge distillation is the ability to compress models without sacrificing accuracy and compromising performance. This is a potential solution for creating deployable models without computation or memory issues. We performed knowledge distillation in the following settings:

\paragraph{Distillation of larger models to MobileNet:}
We performed knowledge distillation from Efficient Net to MobileNet and from ResNet to MobileNet. The distilled MobileNet model has validation accuracies closer to the teacher models despite having almost $7$ times lesser parameters than Efficient Net and $2.6$ times lesser parameters than ResNet. When compared to the teacher Efficient Net, the distilled MobileNet model has a validation accuracy gain of $2.68$\%  with $10$\% labels and a reduction of just $4.09$\% with $25$\% labels. Distillation from ResNet also showed validation accuracies closer to the teacher for $10$\% labels and a small drop in accuracy for $25$\% and $50$\% labels. Given the differences in parameters between the two models, this is rather impressive. The distilled MobileNet model consistently outperforms the MobileNet model that is trained without distillation, indicating that even with fewer parameters, performance can be enhanced by transferring information from a larger model. The results of distillation from Efficient Net to MobileNet and ResNet to MobileNet can be seen in Figure \ref{fig:knoeledge distill} (a) and (b) respectively.

\paragraph{Distillation of Efficient Net to ResNet:}
The distillation of Efficient Net to ResNet showed remarkable results in terms of gain in validation accuracy of the student. Despite lowering the number of parameters to almost half, ResNet's validation accuracy increased by $6.99$\% compared to Efficient Net's accuracy with $10$\% labels, $2.68$\% with $25$\% labels and by $0.87$\% with $50$\% labels. These results are further summarized in Figure \ref{fig:knoeledge distill} (c). The performance is better than the distillation of Efficient Net to MobileNet because of lesser difference in the number of parameters. 

It is interesting that, while the validation accuracies of distilled and undistilled ResNet are identical, the distilled ResNet has a lower training accuracy, which makes it better at generalization.

\subsection{Distillation in supervised setup}
We also carried out the above-stated experiments with fully supervised training. The improvements were not as significant as they were for less labeled data. One reason might be that while using distillation in a semi-supervised setting, KL Divergence Loss is calculated over additional number of datapoints which are unlabeled, thus resulting in increase in performance.

\section{Conclusion}
\label{section: conclusion}
We demonstrate in our study that by distilling semi-supervised models with KL Divergence Loss, we can easily improve their generalization. It has also been shown that a semi-supervised model may be compressed into a smaller model with comparable validation accuracy but greater generalization via knowledge distillation. To summarise, if faced with the challenge of limited labels and memory for deployment, distillation can be a simple method to overcome the problem and make the model deployable. It could also be used as a general practice when performing semi-supervised learning as it is relatively easy to implement with few or no major downsides.

\section{Acknowledgement}
We would like to thank Research Society Manipal for their valuable inputs and research guidance.

{\small
\bibliographystyle{plain}
\bibliography{distillation}
}

\end{document}